\documentclass{article} 
\usepackage{iclr2027_conference,times}
\usepackage{xspace}
\usepackage{pifont}

\usepackage{amsmath,amsfonts,bm}

\def\eqref#1{equation~\ref{#1}}

\def\1{\bm{1}}

\DeclareMathAlphabet{\mathsfit}{\encodingdefault}{\sfdefault}{m}{sl}
\SetMathAlphabet{\mathsfit}{bold}{\encodingdefault}{\sfdefault}{bx}{n}

\definecolor{iccvblue}{rgb}{0.21,0.49,0.74}
\usepackage[colorlinks,linkcolor=red,citecolor=iccvblue]{hyperref}

\usepackage{url}
\usepackage{wrapfig}
\usepackage{graphicx}
\usepackage{caption}
\usepackage{booktabs}
\usepackage{xcolor}
\definecolor{promptblue}{RGB}{24, 74, 117}
\usepackage{pifont}
\usepackage{multirow}
\usepackage[table]{xcolor} 
\usepackage[accsupp]{axessibility}  
\usepackage{amssymb}
\usepackage{arydshln}

\newcommand{\eg}{\textit{e.g.}}
\newcommand{\ie}{\textit{i.e.}}

\definecolor{mygray}{gray}{0.95}

\title{Exploiting Target Knowledge from MLLMs \\for Robust Few-Shot Segmentation}

\author{Yijun Hu$^{1,2}$ \;\;\; Heng Fan$^{3\dagger}$ \;\;\; Libo Zhang$^{2\dagger}$ \\
$^{1}$UCAS \;\;\;
$^{2}$ISCAS \;\;\; 
$^{3}$UNT \\
}

\iclrfinalcopy 
\begin{document}

{
  \renewcommand{\thefootnote}{} 
  \footnotetext{
    $^\dagger$Equal advising and corresponding authors
  }
}
\def\thefootnote{\arabic{footnote}}

\maketitle
\fancyhead{} 
\renewcommand{\headrulewidth}{0pt} 

\begin{abstract}
  Few-shot segmentation (FSS) aims to segment unseen object categories with a few (\eg, one or five) labeled examples, enabling efficient adaptation to novel classes. Conventional models typically rely on appearance-based visual matching between support and query images for segmentation. While straightforward, these methods often struggle to handle significant appearance discrepancies and occlusions in the query image due to insufficient target knowledge. To mitigate this, we introduce a novel framework that mines target knowledge using the strong reasoning capacity of Multimodal Large Language Models (MLLMs) and employs it to enhance FSS. Specifically, building on SAM 2, our method, named MK-FSS, exploits two forms of complementary knowledge derived from a query image by an MLLM for FSS, including spatial knowledge, which provides a spatial prior indicating the potential target location, and semantic knowledge, which describes the target using text. The spatial knowledge is first encoded into a memory representation, and then resulting memory is integrated with the support-guided memory feature from query image through a carefully designed dual-memory debate-fusion (DMDF) module, yielding a more robust target memory feature. In parallel, the semantic knowledge is encoded into the textual feature, which is fused with multi-scale query features via a progressive cross-modal prompt generator (PCPG), producing a target-aware multimodal prompt for segmentation. Working together, the dual-memory feature and the multimodal prompt provide a comprehensive representation of the target, enabling more robust segmentation. In our extensive experiments, MK-FSS shows promising results and largely surpasses existing methods. Code will be released.

\end{abstract}

\section{Introduction}

\label{sec:intro}
Few-shot segmentation (FSS) aims at segmenting novel object classes using a few labeled examples. It is of great practical importance in scenarios where obtaining the pixel-level annotations is time-consuming and labor-intensive, \eg, medical imaging, remote sensing, and autonomous driving. By enabling efficient adaptation to untrained categories, FSS provides a promising solution for scalable and data-efficient visual understanding and received extensive attention in the past decade~\citep{pixel5,proto1,defss,pixel2,text3,abcnet}.

Existing FSS approaches~\citep{ocnet,pahnet,defss} often follow a support-guided segmentation paradigm, where the information extracted from a limited number of labeled support images is leveraged to identify and segment the target in a query image. Despite considerable progress, their performance remains limited in challenging scenarios, such as those involving large appearance variations, occlusions, or visually similar distractors. A key limitation is that using only a small number of support images is insufficient to describe the target across diverse conditions. To mitigate this, recent studies have therefore explored incorporating additional knowledge using visual foundation models~\citep{fss-sam,gf-sam} or class-level textual descriptions derived from category names~\citep{text0,text1,text2} to complement the cues from support images. While these efforts demonstrate improvements, they still suffer from limited exploitation of target-specific information, resulting in suboptimal performance in complex scenarios. This naturally raises a question: \emph{How can we obtain more discriminative and robust target-specific information to improve few-shot segmentation?}

Recently, Multimodal Large Language Models (MLLMs)~\citep{qwen3-vl,comanici2025gemini} have exhibited strong capabilities in visual understanding and reasoning, providing a promising way to extract rich target-specific knowledge directly from the given query image. Drawing inspiration from this, we propose a novel framework that mines target-specific knowledge from MLLMs and leverages such knowledge to enhance FSS. Our method, termed \emph{\textbf{MK-FSS}}, builds upon SAM 2~\citep{ravi2024sam} and exploits two complementary forms of MLLM-derived target-specific knowledge from the query image, including spatial knowledge and semantic knowledge, for FSS. Specifically, the spatial knowledge offers a spatial prior indicating the potential target location, while the semantic knowledge describes the target object through textual information. To obtain a more discriminative target representation, we first encode  the spatial knowledge into a memory representation, and then integrate it with the support-guided memory feature from query image through a carefully designed dual-memory debate-fusion (DMDF) module. This module regards the two memory features as two complementary agents, which exchange, verify, and refine information through a structured debate-verification-integration process, thereby yielding a more robust target memory feature. Meanwhile, the semantic knowledge is first encoded into the textual feature, which is then fused with multi-scale query features via a progressive cross-modal prompt generator (PCPG). By explicitly injecting high-level linguistic semantics into visual space, PCPG progressively integrates textual and visual features to produce a target-aware multimodal prompt for segmentation. Working together, the dual-memory feature and the multimodal prompt provide a comprehensive target representation, enabling accurate and robust segmentation. To validate the effectiveness of our MK-FSS, we conduct experiments on PASCAL-$5^i$~\citep{proto4} and COCO-$20^i$~\citep{coco20}. MK-FSS achieves new state-of-the-art performance on both benchmarks, evidencing its efficacy.

We note several recent works~\citep{llafs,dsv-lfs} have attempted to exploring LLMs for improving FSS. MK-FSS \emph{\textbf{differs from}} these methods in two key aspects. First, the type of knowledge exploited for FSS is different. Existing methods either leverage LLM-generated polygon predictions~\citep{llafs} or category-level textual descriptions~\citep{dsv-lfs}, whereas our method extracts target-specific spatial and semantic knowledge directly from the query image, enabling the exploration of more discriminative information for segmentation. Second, the way in which LLM-derived knowledge is utilized is also different. Current approaches mainly use LLM-derived knowledge as direct segmentation guidance~\citep{llafs} or combine it with visual prompts~\citep{dsv-lfs}, while MK-FSS integrates MLLM-derived spatial and semantic knowledge for target representation learning, which achieves superior performance.

In summary, our contributions are summarized as follows: \ding{171}	We propose MK-FSS, a novel framework that exploits complementary MLLM-derived spatial and semantic knowledge for FSS; \ding{170} We present a dual-memory debate-fusion (DMDF) module that integrates memory features derived from spatial knowledge and support information, yielding a more robust dual-memory representation. \ding{168} We propose a progressive cross-modal prompt generator (PCPG) that fuses textual feature encoded from semantic knowledge with multi-scale query features to produce a multimodal prompt for FSS. \ding{169} In extensive experiments, MK-FSS shows promising results, evidencing its effectiveness.

\section{Related Work}

\textbf{Few-Shot Segmentation.} Few-shot segmentation (FSS)~\citep{proto4} aims to segment novel object categories using only a few annotated samples. Current approaches mainly follow prototype-based~\citep{proto1,proto2,proto3,proto5,proto6,pfenet} or feature-matching-based~\citep{pixel1,pixel2,pixel3,pixel4} paradigms. The former represents support information with prototypes to guide the segmentation of query images, while the latter exploits pixel-level support-query correspondences to achieve segmentation. Despite their effectiveness, both rely heavily on the limited support samples and may struggle with complex cases such as occlusions and visually similar distractors. Recent works incorporate foundation models~\citep{gf-sam,fss-sam,foundation3,liu2023matcher} or textual cues derived from category names~\citep{text1,text2,text3,text4} to provide additional target knowledge for FSS. Yet, such auxiliary information remains limited in resolving target ambiguity. More recently, Large Language Models (LLMs) have been explored to provide spatial localization cues~\citep{llafs} or generate category-level textual descriptions~\citep{dsv-lfs} for FSS. Despite their improvements, these methods do not fully exploit query-specific target information. Different from them, MK-FSS explicitly harnesses advanced reasoning capabilities of Multimodal Large Language Models (MLLMs) to derive complementary spatial and semantic knowledge from the query image, enabling more reliable target segmentation in complex scenes.

\noindent\textbf{Multimodal Large Language Models.} Multimodal Large Language Models (MLLMs) have shown remarkable capabilities in multimodal processing and reasoning in recent years~\citep{gpt3, llava,qwen3-vl,comanici2025gemini}. Given their extensive internal knowledge bases, numerous downstream visual tasks have leveraged these powerful models for auxiliary guidance~\citep{llm4sgg,VisDiff,lisa,jiang2025multimodal,industrial,LG-SOR}. For example, the work of SEAL~\citep{seal} integrates an MLLM to provide informative visual cues for detection, while DeepEeyes~\citep{Deepeyes} leverages the inherent grounding capability of MLLMs to identify potential target regions. Different from these methods, our proposed MK-FSS specifically employ MLLM to generate spatial and semantic target-relevant knowledge from the query image and incorporate them into FSS through two dedicated modules.

\section{Our Approach}

\textbf{Problem Definition}
FSS aims to segment targets in a query image $I_q$ using only $K$ (typically $K=1$ or 5) annotated support samples $\{(I_s^i, M_s^i)\}_{i=1}^K$ from the same semantic category, where $I_s^i$ and $M_s^i$ denote the $i$-th support image and its corresponding mask. Following the standard episodic training protocol, each episode samples a target class $\xi$ and contains a support set $S$ and a query pair $(I_q, M_q)$, where $M_q$ is the groundtruth mask in $I_q$ and used only for training supervision. The model is trained on base classes $\xi_{\text{train}}$ and evaluated on disjoint novel classes $\xi_{\text{test}}$, with $\xi_{\text{train}} \cap \xi_{\text{test}} = \emptyset$. Following prior LLM/MLLM-based FSS methods~\citep{dsv-lfs,llafs}, we also use class label $\xi$ to elicit target-related knowledge from the MLLM. Since the class label is readily available in standard FSS benchmarks, this process does not require any additional manual annotations.

\begin{figure}[!t]
  \centering
    \includegraphics[width=\linewidth]{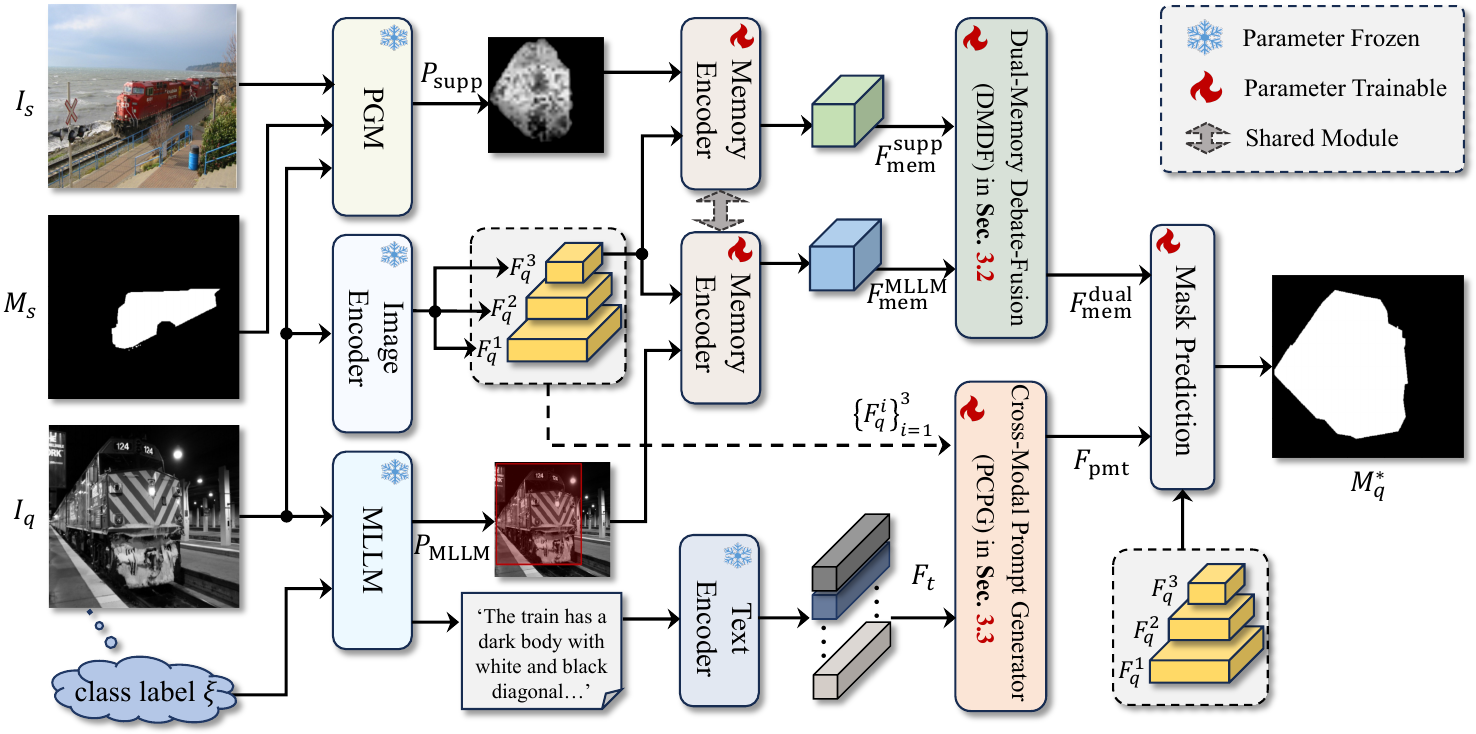}\vspace{2mm}
  \caption{Overview of MK-FSS that exploits complementary spatial and semantic knowledge derived from a query image by an MLLM for few-shot segmentation.}
  \label{fig:model}
\end{figure}

\textbf{Overview.} In this work, we introduce MK-FSS, which exploits MLLM-derived target knowledge to improve FSS. As in Fig.~\ref{fig:model}, building upon SAM 2~\citep{ravi2024sam}, MK-FSS first generates the target-specific knowledge from the query image using MLLM, and then leverages such knowledge to enhance target representation (\S\ref{sec:preparation}). Specifically, we first encode the spatial knowledge into a memory representation, and then integrate it with the support-guided memory feature
derived from the query image through our  DMDF, yielding a more robust dual-memory feature (\S\ref{sec:DMDF}). Meanwhile, the semantic knowledge is encoded into a textual representation, which is progressively injected into query features by PCPG to produce a multimodal prompt (\S\ref{sec:PCPG}). The resulting dual-memory feature and the multimodal prompt are jointly used for mask prediction, enabling more robust segmentation.

\subsection{Our MK-FSS Framework}
\label{sec:preparation}

In the following, we detail our MK-FSS by illustrating how target-specific knowledge derived from MLLM is leveraged to enhance few-shot segmentation.

\subsubsection{Target Knowledge Generation}
\label{knowledge}

Our MK-FSS exploits two forms of target knowledge from an MLLM, including spatial knowledge and semantic knowledge, to improve FSS. The former provides spatial cues about the target location in the query image, while the latter captures high-level semantic information of the target. Together, they offer complementary information for robust few-shot segmentation.

\textbf{Spatial Knowledge.} Specifically, to generate spatial knowledge, we send the query image $I_q$ and class label $\xi$ to MLLM, and prompt it to identify all potential target regions and output their bounding box coordinates. The predicted boxes are then post-processed and mapped onto the query image to construct an MLLM-derived binary spatial prior $P_{\text{MLLM}} \in \{0,1\}^{H \times W}$, where $H$ and $W$ denote the height and width of the query image, highlighting the potential target regions in the query image. Let $\{b_i\}_{i=1}^{n}$ denote the $n$ bounding boxes predicted by MLLM. Then, $P_{\text{MLLM}}$ is defined as
\begin{equation}
\setlength{\abovedisplayskip}{5pt} 
\setlength{\belowdisplayskip}{5pt}
P_\text{MLLM}(x, y) =
\begin{cases}
1, & \text{if } (x, y) \in \bigcup_{i=1}^{n} b_n, \\
0, & \text{otherwise,}
\end{cases}
\end{equation}
where $(x, y)$ denotes a pixel location in the query image.

\textbf{Semantic Knowledge.} For semantic knowledge generation, we further prompt the MLLM with $I_q$ and $\xi$ to generate a textual description of the target, including its appearance and contextual cues. The generated description is encoded by a text encoder to produce a  holistic sentence embedding $F_s \in \mathbb{R}^{1 \times C}$ and word-level embedding $F_w \in \mathbb{R}^{L \times C}$, where $L$ is the embedding sequence length and $C$ denotes the feature dimension. These two representations are concatenated to form a unified textual feature $F_t=\mathtt{Concat}(F_s,F_w) \in \mathbb{R}^{(L+1) \times C}$ for subsequent multimodal prompt generation.

Due to space limitations, please refer to the \textbf{\emph{appendix}} for prompt templates used by the MLLM and representative examples of the generated spatial and semantic knowledge from the query image.

\subsubsection{Few-Shot Segmentation with Target Knowledge}

Given the target knowledge generated above, MK-FSS leverages it to improve FSS. Specifically, the spatial knowledge is encoded into the memory feature and integrated with a support-guided memory to construct a robust target memory, while the semantic knowledge is encoded into a textual feature and fused with multi-scale query features to generate a multimodal prompt for FSS.

Specifically, given a support set $\{(I_s^i, M_s^i)\}_{i=1}^K$ and the query image $I_q$, we first use a prior generation module (PGM), following~\citep{fss-sam}, to generate the support-guided spatial prior. For clarity, we take the one-shot setting, \ie, $K=1$, as the example for illustration. Concretely, within PGM, a vision foundation model $\Phi_{v}(\cdot)$~\citep{oquab2023dinov2} is first used to extract query feature $\tilde{F}_q = \Phi_{v}(I_q)$ and support feature $\tilde{F}_s = \Phi_{v}(I_s)$. Afterwards, based on the support mask $M_s$, $\tilde{F}_s$ is compressed into foreground prototype $F_{\text{fg}}$ and background prototype $F_{\text{bg}}$ using global average pooling via $F_{\text{fg}}=\mathtt{GAP}(\tilde{F}_s \odot M_s)$ and $F_{\text{bg}}=\mathtt{GAP}(\tilde{F}_s \odot (1-M_s))$, where $\mathtt{GAP}(\cdot)$ denotes global average pooling and $\odot$ is element-wise multiplication. Finally, we compute the cosine similarity between the query feature $\tilde{F}_q$ and the two prototypes $F_{fg}$ and $F_{bg}$, and obtain a non-negative support-guided spatial prior $P_{\text{supp}}$ by subtracting the background similarity from the foreground similarity, as follows, 
\begin{equation}
\setlength{\abovedisplayskip}{5pt} 
\setlength{\belowdisplayskip}{5pt}
P_{\text{supp}} = \mathtt{max}(\mathtt{CosSim}(F_{\text{fg}}, \tilde{F}_q) - \mathtt{CosSim}(F_{\text{bg}}, \tilde{F}_q), 0)
\end{equation}
where $\mathtt{CosSim}(\cdot,\cdot)$ is the cosine similarity function. Due to space limitation, the detailed architecture of PGM~\citep{fss-sam} is shown in the \emph{\textbf{appendix}}. 

Once obtaining the support-guided spatial prior $P_{\text{supp}}$, we use it to generate the support-guided memory feature from the query image using memory encoder of SAM 2~\citep{ravi2024sam}. Specifically, the query image $I_q$ is first sent to the image encoder of SAM 2 to obtain multi-scale query features $\{F_q^1, F_q^2, F_q^3\}$, where $F_q^1$ to $F_q^3$ correspond to progressively deeper feature levels. Then, we use $F_q^3$ and $P_{\text{supp}}$ to generate the support-guided memory $F_{\text{mem}}^{\text{supp}}$, as follows,
\begin{equation}
\setlength{\abovedisplayskip}{5pt} 
\setlength{\belowdisplayskip}{5pt}
F_{\text{mem}}^{\text{supp}} = \mathtt{MemEnc}(F_q^3,P_{\text{supp}})
\end{equation}
where $\mathtt{MemEnc}(\cdot,\cdot)$ is the memory encoder
of SAM 2. Because of large appearance variations, using $F_{\text{mem}}^{\text{supp}}$ alone may be insufficient to accurately segment the target due to lacking target knowledge. To mitigate this, we introduce an additional memory derived from the spatial knowledge provided by the MLLM and integrate it with $F_{\text{mem}}^{s}$ for a more reliable memory. Specifically, given the MLLM-derived spatial prior $P_{\text{MLLM}}$ and $F_q^3$, we generate the MLLM-derived memory $F_{\text{mem}}^{\text{MLLM}}$ as follows,
\begin{equation}
\setlength{\abovedisplayskip}{5pt} 
\setlength{\belowdisplayskip}{5pt}
F_{\text{mem}}^{\text{MLLM}} = \mathtt{MemEnc}(F_q^3,P_{\text{MLLM}})
\end{equation}
After obtaining $F_{\text{mem}}^{\text{supp}}$ and $F_{\text{mem}}^{\text{MLLM}}$, we integrate them with DMDF (explained in Sec.~\ref{sec:DMDF}) via 
\begin{equation}
\setlength{\abovedisplayskip}{5pt} 
\setlength{\belowdisplayskip}{5pt}
F_{\text{mem}}^{\text{dual}} = \mathtt{DMDF}(F_{\text{mem}}^{\text{MLLM}}, F_{\text{mem}}^{\text{supp}})
\end{equation}
where $F_{\text{mem}}^{\text{dual}}$ is the resulting dual-memory target feature.

Besides the spatial knowledge, we further employ the semantic knowledge derived from the MLLM to provide high-level target features for FSS, which is beneficial for disambiguating visually similar regions in the query image. Specifically, given multi-scale query features $\{F_q^1, F_q^2, F_q^3\}$ and textual feature $F_t$ encoded from the semantic knowledge (see Sec.~\ref{knowledge}), we fuse them via PCPG (explained in Sec.~\ref{sec:PCPG}) to generate a multimodal prompt for segmentation as follows,
\begin{equation}
\setlength{\abovedisplayskip}{5pt} 
\setlength{\belowdisplayskip}{5pt}
F_{\text{pmt}} = \mathtt{PCPG}(F_q^1, F_q^2, F_q^3, F_t) 
\end{equation}
where $F_{\text{pmt}}$ is the learned target-aware multimodal prompt for segmentation.

Finally, we combine the dual-memory feature $F_{\text{mem}}^{\text{dual}}$ and multimodal prompt $F_{\text{pmt}}$ to jointly guide the mask decoder for final target segmentation, as follows,
\begin{equation}
\setlength{\abovedisplayskip}{5pt} 
\setlength{\belowdisplayskip}{5pt}
M_q^* =
\mathtt{MaskDec}
(
F_q^{1}, F_q^{2}, \hat{F}_q^3
)
\quad
\hat{F}_q^3 =
\mathtt{MemAtt}
(
F_q^3 + F_{\text{pmt}},
F_{\text{mem}}^{\text{dual}}
)
\end{equation}
where $\mathtt{MemAtt}(\cdot,\cdot)$ and $\mathtt{MaskDec}(\cdot, \cdot, \cdot)$ are the memory attention and mask decoder of SAM 2, and $M_q^*$ is the predicted segmentation mask in the query image $I_q$.

\subsection{Dual-Memory Debate-Fusion}
\label{sec:DMDF}

\setlength{\columnsep}{10pt}%
\setlength\intextsep{0pt}
\begin{wrapfigure}{r}{0.58\textwidth}
\centering
\includegraphics[width=0.58\textwidth]{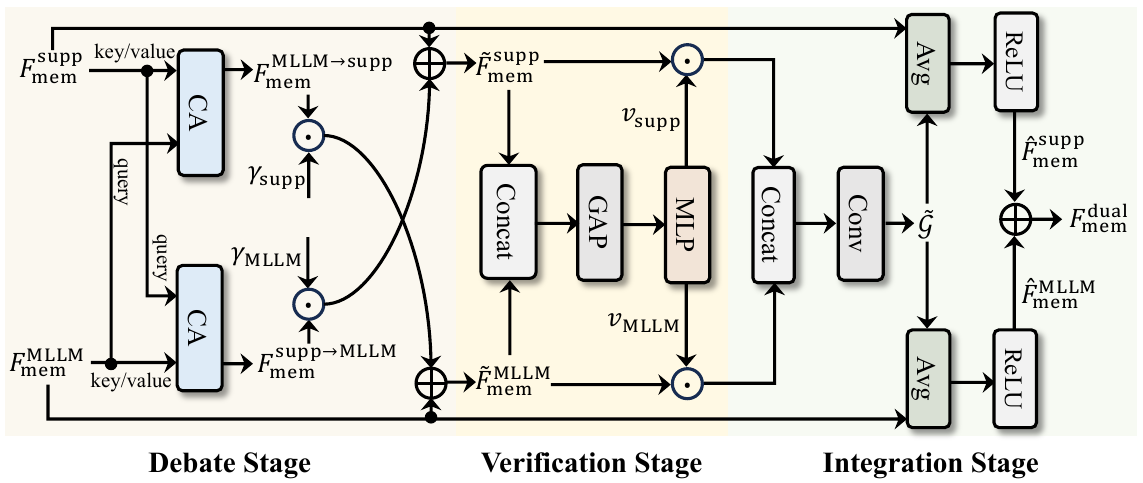}
\caption{Overview of DMDF.}
\vspace{2mm}
\label{fig:memory_fusion}
\end{wrapfigure}
The recent LLM systems have applied the Multi-Agent Debate (MAD) mechanism~\citep{debate1, chan2024chateval} to boost the reasoning reliability. In these systems, multiple LLM agents discuss the shared question and update their answers based on peer responses, yielding more robust reasoning and improved outcomes. Inspired by this, we introduce a novel dual-memory debate-fusion (DMDF) module that mutually enhances the support-guided and MLLM-derived memories, producing a more robust dual-memory feature. Analogous to the MAD process in LLM systems, DMDF operates in three stages, including debate, verification and integration, as described in the following.

\textbf{Debate Stage.} In debate stage, each memory feature attends to the other to extract complementary information, enabling bilateral information exchange analogous to the debate between two agents. Specifically, given the support-guided memory feature $F_{\text{mem}}^{\text{supp}}$ and the MLLM-derived memory feature $F_{\text{mem}}^{\text{MLLM}}$, we apply a bidirectional cross-attention mechanism to facilitate mutual feature refinement. As illustrated in Fig.~\ref{fig:memory_fusion}, this process can be expressed as follows,
\begin{equation}
\setlength{\abovedisplayskip}{5pt} 
\setlength{\belowdisplayskip}{5pt}
F_{\text{mem}}^{\text{MLLM} \rightarrow \text{supp}} = \mathtt{CA}(F_{\text{mem}}^{\text{supp}}, F_{\text{mem}}^{\text{MLLM}}), \quad
F_{\text{mem}}^{\text{supp} \rightarrow \text{MLLM}} = \mathtt{CA}(F_{\text{mem}}^{\text{MLLM}}, F_{\text{mem}}^{\text{supp}})
\end{equation}
where $F_{\text{mem}}^{\text{MLLM} \rightarrow \text{supp}}$ and $F_{\text{mem}}^{\text{supp} \rightarrow \text{MLLM}}$ are refined memory features, and $\mathtt{CA}(\textbf{z},\textbf{u})$ represents the cross-attention operation with \textbf{z} generating query and \textbf{u} key/value. The resulting refined features are then fused with the original memory features to incorporate complementary information from each other, yielding the updated memory features as follows,
\begin{equation}
\tilde{F}_{\text{mem}}^{\text{supp}} = F_{\text{mem}}^{\text{supp}} + \gamma_{\text{supp}} \cdot F_{\text{mem}}^{\text{MLLM} \rightarrow \text{supp}} \quad
\tilde{F}_{\text{mem}}^{\text{MLLM}} = F_{\text{mem}}^{\text{MLLM}} + \gamma_{\text{MLLM}} \cdot F_{\text{mem}}^{\text{supp} \rightarrow \text{MLLM}}
\end{equation}
where $\tilde{F}_{\text{mem}}^{\text{supp}}$ and $\tilde{F}_{\text{mem}}^{\text{MLLM}}$ denotes updated memory features, and $\gamma_{\text{supp}}$ and $\gamma_{\text{MLLM}}$ are learnable scaling factors that control the memory fusion in the debate stage. 

\textbf{Verification Stage.} After obtaining the updated memories, we estimate channel-wise importance weights for two memories, enabling subsequent fusion to emphasize more informative cues. Specifically, the two memory features are globally pooled and concatenated along the channel dimension. Then, the concatenated feature is fed into a multilayer perceptron (MLP) to generate the reliability weights for the two memories, as follows,
\begin{equation}
v_{\text{supp}}, v_{\text{MLLM}} =
\mathtt{MLP}(
\mathtt{Concat}(\mathtt{GAP}(\tilde{F}_{\text{mem}}^{\text{supp}}),
\mathtt{GAP}(\tilde{F}_{\text{mem}}^{\text{MLLM}}))
)
\end{equation}
where $v_{\text{supp}}$ and $v_{\text{MLLM}}$ are reliability weights of the support-guided and MLLM-derived memories.

\textbf{Integration Stage.} In the integration stage, we fuse the updated memories according to their reliability weights to produce the final dual-memory feature. Specifically, each updated memory is first weighted by its corresponding reliability weight, and the weighted memories are concatenated and passed through a convolution layer to obtain a combined representation $\tilde{\mathcal{G}}$ via
\begin{equation}
\tilde{\mathcal{G}} =
\mathtt{Conv}(
\mathtt{Concat}(
\tilde{F}_{\text{mem}}^{\text{supp}} \odot v_{\text{supp}},
\tilde{F}_{\text{mem}}^{\text{MLLM}} \odot v_{\text{MLLM}}
)
)
\end{equation}
where $\mathtt{Conv}(\cdot)$ denotes the convolution operation. After this, the combined representation $\tilde{\mathcal{G}}$ is fused with each original memory feature by averaging, followed by a ReLU activation to obtain the final memory features, as follows,
\begin{equation}
\hat{F}_{\text{mem}}^{\text{supp}} = \mathtt{ReLU}(\mathtt{Avg}(\tilde{\mathcal{G}}, F_{\text{mem}}^{\text{supp}})) \quad
\hat{F}_{\text{mem}}^{\text{MLLM}} = \mathtt{ReLU}(\mathtt{Avg}(\tilde{\mathcal{G}}, F_{\text{mem}}^{\text{MLLM}})) \\
\end{equation}
where $\mathtt{Avg}(\cdot)$ is the averaging operation and $\mathtt{ReLU}(\cdot)$ denotes the ReLU activation. The final dual-memory target feature $F_{\text{mem}}^{\text{dual}}$ is then obtained via
\begin{equation}
F_{\text{mem}}^{\text{dual}} = \hat{F}_{\text{mem}}^{\text{supp}} + \hat{F}_{\text{mem}}^{\text{MLLM}}
\end{equation}
By effectively incorporating support-guided and MLLM-derived memories, $F_{\text{mem}}^{\text{dual}}$ provides a more reliable target representation for target segmentation.

\subsection{Progressive Cross-modal Prompt Generator}
\label{sec:PCPG}

\setlength{\columnsep}{10pt}%
\setlength\intextsep{0pt}
\begin{wrapfigure}{r}{0.465\textwidth}
\centering
\includegraphics[width=0.465\textwidth]{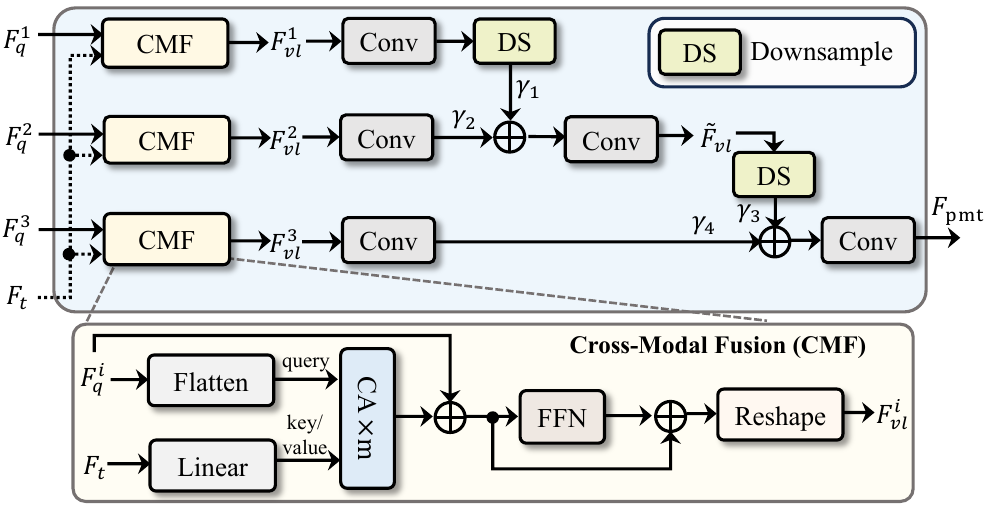}
\caption{Overview of PCPG.}
\vspace{1mm}
\label{fig:cross_modal}
\end{wrapfigure}
To further exploit the semantic knowledge, we propose a progressive cross-modal prompt generator (PCPG) that fuses multi-scale query features with the textual feature to produce a multimodal prompt for segmentation. As illustrated in Fig.~\ref{fig:cross_modal}, PCPG first applies a cross-modal fusion (CMF) module (as described later) to each query feature $F_{q}^{i}$, where $i\in \{1,2,3\}$, together with the textual feature $F_t$, to generate cross-modal features $F_{vl}^{i}$ as follows,
\begin{equation}
\setlength{\abovedisplayskip}{5pt} 
\setlength{\belowdisplayskip}{5pt}
F_{vl}^{i} = \mathtt{CMF}(F_q^i,F_t) \;\;\; i\in\{1,2,3\}
\end{equation}
The resulting multi-scale cross-modal features $F_{vl}^{i}$ are then progressively aggregated from the shallow to deep levels. Specifically, $F_{vl}^{1}$ and $F_{vl}^{2}$ are first fed into separate convolutional layers, after which the former is downsampled and combined with the latter using learnable weights. The fused feature is further processed by a convolutional layer to obtain the intermediate representation $\tilde F_{vl}$. Afterwards, $\tilde F_{vl}$ is downsampled and combined with the convolutionally processed $F_{vl}^{3}$ using learnable weights, followed by another convolutional layer to generate the multimodal prompt. Mathematically, this process can be expressed as follows,
\begin{equation}
\setlength{\abovedisplayskip}{5pt} 
\setlength{\belowdisplayskip}{5pt}
\begin{aligned}
\tilde F_{vl}
&= \mathtt{Conv}(
\mathtt{DS}\left(\gamma_{1} \cdot \mathtt{Conv}(F_{vl}^{1})\right)
+ \gamma_{2} \cdot \mathtt{Conv}(F_{vl}^{2})
) \\
F_{\text{pmt}}
&= \mathtt{Conv}(
\gamma_{3} \cdot \mathtt{DS}(\tilde F_{vl})
+ \gamma_{4} \cdot \mathtt{Conv}(F_{vl}^{3})
)
\end{aligned}
\end{equation}
where $F_{\text{pmt}}$ is the generated multimodal prompt feature, $\{\gamma_{i}\}_{i=1}^{4}$ are learnable parameters, and $\mathtt{DS}(\cdot)$ represents the downsampling operation.

\textbf{Cross-Modal Fusion.}
The cross-modal fusion (CMF) module aims to fuse the textual feature $F_t$ with the query feature $F_q^i$. Specifically, $F_q^i$ is first flattened into visual tokens, while $F_t$ is linearly projected to match its channel dimension. We then apply $m$ ($m$ is empirically set to 2) consecutive cross-attention layers to enable interaction between $F_q^i$ and $F_t$, followed by a residual connect. The resulting fused feature is further processed by an FFN with another residual connection and reshaped to obtain the cross-modal feature $F_{vl}^i$. Fig.~\ref{fig:cross_modal} shows the architecture of CMF.

\subsection{$K$-shot Extension and Optimization}

\textbf{Extending to $K$-shot Setting.} In $K$-shot FSS, each support sample is applied to generate a support-guided prior mask. We average all these masks to obtain the final support-guided spatial prior $P_{\text{supp}}$. Once $P_{\text{supp}}$ is obtained, the subsequent FSS procedure is identical to that in the one-shot setting.

\textbf{Optimization.} Given a query image and a support set, MK-FSS predicts a foreground segmentation mask $M_q^{*}$. During the training process, the model is optimized using the standard Dice loss~\citep{milletari2016v} between $M_q^{*}$ and the target groundtruth mask.

\section{Experiments}

\textbf{Implementation.} MK-FSS is implemented in Python using PyTorch~\citep{paszke2019pytorch} and built upon the pretrained SAM 2-small model~\citep{ravi2024sam}. The memory attention and mask decoder modules are fine-tuned, while the newly introduced modules are trained from scratch. In MK-FSS, we employ Qwen3-VL-8B-Instruct~\citep{qwen3-vl} as the MLLM. To preserve localization cues, we post-process the bounding boxes predicted by the MLLM. Specifically, the invalid outputs, such as empty or malformed boxes, are replaced with a box that covers the entire image; otherwise, the predicted boxes are utilized as the spatial prior. We employ RoBERTa~\citep{liu2019roberta} as the text encoder. For PGM, DINOv2-B~\citep{oquab2023dinov2} is applied for visual feature extraction. During training, images are randomly cropped to $512\times512$, and predictions are resized to their original resolution for evaluation. The model is trained using AdamW~\citep{adamw} with an initial learning rate of 0.001 for 150 and 100 epochs on PASCAL-${5^i}$ and COCO-${20^i}$, respectively. All models are trained on three NVIDIA A100 (40G) GPUs and tested on a single A100 (40G) GPU.

\subsection{Datasets and Evaluation Metrics.}

\textbf{Datasets.} We evaluate MK-FSS on PASCAL-$5^i$~\citep{proto4} and COCO-$20^i$~\citep{coco20}. PASCAL-$5^i$ includes 20 classes, built upon the PASCAL VOC 2012~\citep{pascalVOC} with additional annotations from SDS~\citep{SDS}. COCO-$20^i$~\citep{MSCOCO} is larger in scale and comprises 80 classes. Both datasets are divided into 4 disjoint folds for cross-validation, with each fold containing 5 classes for
PASCAL-$5^i$ and 20 classes for COCO-$20^i$. In each iteration, 3 folds are used for training, and the remaining fold is used for testing. Following existing works, we randomly sample 1,000 episodes for testing by default. Due to limited space, the results for error bars evaluation and different number of testing episodes are included in \emph{\textbf{appendix}}.

\textbf{Evaluation Metrics.} Following prior works, we employ mean intersection-over-union (mIoU) and foreground-background IoU (FB-IoU) as evaluation metrics. Specifically, mIoU averages IoU across foreground classes within each fold, while FB-IoU treats all foreground classes as a single category and measures binary foreground-background segmentation performance.

\renewcommand{\arraystretch}{1.1}
\begin{table*}[t]
\setlength{\tabcolsep}{6pt}
\centering
\caption{Comparisons with state-of-the-arts on PASCAL-5$^i$. Best results are highlighted in \textbf{bold}.}
\resizebox{\linewidth}{!}{  
\begin{tabular}{rcccccccccccc}
\specialrule{1.5pt}{0pt}{0pt}
\multirow{2}{*}{\textbf{Method}} & \multicolumn{6}{c}{\textbf{1-shot}} & \multicolumn{6}{c}{\textbf{5-shot}} \\
\cmidrule(lr){2-7} \cmidrule(lr){8-13}
& $\mathbf{5^0}$ & $\mathbf{5^1}$ & $\mathbf{5^2}$ & $\mathbf{5^3}$ & \textbf{Mean} & \textbf{FB-IoU} & $\mathbf{5^0}$ & $\mathbf{5^1}$ & $\mathbf{5^2}$ & $\mathbf{5^3}$ & \textbf{Mean} & \textbf{FB-IoU} \\
\hline\hline
\rowcolor[HTML]{eaf4fc}  
\multicolumn{13}{c}{\textbf{Non-LLM/MLLM-Based Few-Shot Segmentation}} \\
MSI~\citep{msi} \textsubscript{[ICCV'23]} & 73.1 & 73.9 & 64.7 & 68.8 & 70.1 &82.3 &73.6 & 76.1 & 68.0 & 71.3 & 72.2 &82.3\\
PI-CLIP~\citep{text3} \textsubscript{[CVPR'24]} & 76.4 & 83.5 & 74.7 & 72.8 & 76.8 &- & 76.7 & 83.8 & 75.2 & 73.2 & 77.2 &-\\
AENet~\citep{aenet} \textsubscript{[ECCV'24]} & 72.2 & 75.5 &  68.5 & 63.1 & 69.8 &80.8 & 74.2 & 76.5 & 74.8 & 70.6 & 74.1&84.5 \\
GF-SAM~\citep{gf-sam} \textsubscript{[NeurIPS'24]} & 71.1 & 75.7 &  69.2 & 73.3 & 72.1 & -& 81.5 & 86.3 & 79.7 & 82.9 & 82.6&- \\
OCNet~\citep{ocnet} \textsubscript{[ICCV'25]} & 73.5 & 75.9 & 71.1 & 64.9 & 71.4 &82.2 & 75.9 & 77.1 & 74.1 & 70.9 & 74.5 &84.7\\
PAHNet~\citep{pahnet} \textsubscript{[ICCV'25]} & 72.1 & 76.4 & 71.0 & 66.6 & 71.5 & 82.4& 75.8 & 78.6 & 76.9 & 73.2 & 76.1 &85.7\\
DeFSS~\citep{defss} \textsubscript{[ICCV'25]} & 73.7 & 76.6 & 70.9 & 65.8 & 71.7 &- &75.0 & 78.6 & 74.7 & 72.2 & 75.1&- \\
FSSAM~\citep{fss-sam} \textsubscript{[ICML'25]} & 81.6 & 84.9 & 81.6 & 76.0 & 81.0 & 89.4& 84.1 & 88.5 & 83.8 & 85.0 & 85.4 &91.9\\
\hline
\rowcolor[HTML]{eaf4fc}  
\multicolumn{13}{c}{\textbf{LLM/MLLM-based Few-Shot Segmentation}} \\
LLaFS~\citep{llafs} \textsubscript{[CVPR'24]} & 74.2 & 78.8 & 72.3 & 68.5 & 73.5 &- & 75.9 & 80.1 & 75.8 & 70.7 & 75.6 &-\\
DSV-LFS~\citep{dsv-lfs} \textsubscript{[CVPR'25]} & 71.7 & 82.0 & 71.2 & 75.0 & 75.0 &- & 72.0 & 82.0 & 71.3 & 75.5 & 75.2 &-\\
MLLM+SAM 2 (our baseline) & 74.7 & 79.2 & 73.5 & 75.8 & 75.8 & 86.9 &  74.7& 79.3 & 72.7 & 75.3 & 75.5 &86.1\\
\rowcolor[HTML]{EFF7E6} 
\textbf{MK-FSS (ours)} & \textbf{85.0} & \textbf{89.3} & \textbf{86.1} & \textbf{87.8} & \textbf{87.1} &\textbf{93.0} & \textbf{85.2} & \textbf{90.8} & \textbf{85.8} & \textbf{88.9} & \textbf{87.7} &\textbf{93.3}\\
\specialrule{1.5pt}{0pt}{0pt}
\end{tabular}
}
\label{tab:main_result_pascal}\vspace{-0mm}
\end{table*}

\begin{table*}[t]
\setlength{\tabcolsep}{6pt}
\centering
\caption{Comparisons with state-of-the-arts on COCO-20$^i$. Best results are highlighted in \textbf{bold}.}
\resizebox{\linewidth}{!}{  
\begin{tabular}{rcccccccccccc}
\specialrule{1.5pt}{0pt}{0pt}
\multirow{2}{*}{\textbf{Method}} & \multicolumn{6}{c}{\textbf{1-shot}} & \multicolumn{6}{c}{\textbf{5-shot}} \\
\cmidrule(lr){2-7} \cmidrule(lr){8-13}
& $\mathbf{20^0}$ & $\mathbf{20^1}$ & $\mathbf{20^2}$ & $\mathbf{20^3}$ & \textbf{Mean} & \textbf{FB-IoU} & $\mathbf{20^0}$ & $\mathbf{20^1}$ & $\mathbf{20^2}$ & $\mathbf{20^3}$ & \textbf{Mean} & \textbf{FB-IoU} \\
\hline\hline
\rowcolor[HTML]{eaf4fc}  
\multicolumn{13}{c}{\textbf{Non-LLM/MLLM-Based Few-Shot Segmentation}} \\
MSI~\citep{msi} \textsubscript{[ICCV'23]} & 44.8 & 54.2 & 52.3 & 48.0 & 49.8 & -& 49.3 & 58.0 & 56.1 & 52.7 & 54.0 &-\\
PI-CLIP~\citep{text3} \textsubscript{[CVPR'24]} & 49.3 & 65.7 & 55.8 & 56.3 & 56.8 &- & 56.4 & 66.2 & 55.9 & 58.0 & 59.1 &-\\
AENet~\citep{aenet} \textsubscript{[ECCV'24]} & 43.1 & 56.0 &  50.3 & 48.4 & 49.4 &73.6& 51.7 & 61.9 & 57.9 & 55.3 & 56.7 &76.5\\
GF-SAM~\citep{gf-sam} \textsubscript{[NeurIPS'24]} & 56.6 & 61.4 &  59.6 & 57.1 & 58.7 &-& 67.1 & 69.4 & 66.0 & 64.8 & 66.8 &-\\
OCNet~\citep{ocnet} \textsubscript{[ICCV'25]} & 45.9 & 56.9 &  52.9 & 50.4 & 51.5 &73.7 & 52.7 & 63.1 & 57.4 & 54.8 & 57.0 &76.8\\
PAHNet~\citep{pahnet} \textsubscript{[ICCV'25]} & 45.2 & 58.2 & 54.4 & 51.2 & 52.3 &75.2& 53.0 & 64.1 & 60.8 & 58.7 & 59.2&78.1 \\
DeFSS~\citep{defss} \textsubscript{[ICCV'25]} & 46.5 & 59.7 & 54.8 & 53.2 & 53.7 &-& 55.1 & 66.0 & 61.3 & 57.5 & 60.0 &-\\
FSSAM~\citep{fss-sam} \textsubscript{[ICML'25]} & 59.9 & 65.6 & 62.1 & 61.6 & 62.3 &77.3& 68.6 & 74.0 & 64.5 & 69.9 & 69.3  &82.9\\
\hline
\rowcolor[HTML]{eaf4fc}  
\multicolumn{13}{c}{\textbf{LLM/MLLM-based Few-Shot Segmentation}} \\
LLaFS~\citep{llafs} \textsubscript{[CVPR'24]} & 47.5 & 58.8 & 56.2 & 53.0 & 53.9 &-& 53.2 & 63.8 & 63.1 & 60.0 & 60.0 &-\\
DSV-LFS~\citep{dsv-lfs} \textsubscript{[CVPR'25]} & 70.0 & 73.3 & 70.7 & 71.3 & 71.3 & -&71.0 & 73.8 & 71.3 & 71.4 & 71.9 &-\\
 
MLLM+SAM 2 (our baseline) & 64.3 & 64.2 & 68.3 & 61.1 & 64.5 &78.7 & 65.9 & 63.3 & 65.8 & 61.6 &64.2  &79.0\\
\rowcolor[HTML]{EFF7E6} 
\textbf{MK-FSS (ours)} & \textbf{73.2} & \textbf{76.6} & \textbf{75.9} & \textbf{71.6} & \textbf{74.3} &\textbf{84.8} & \textbf{75.8} & \textbf{77.1} & \textbf{76.4} & \textbf{74.3} & \textbf{75.9} &\textbf{86.1}\\
\specialrule{1.5pt}{0pt}{0pt}
\end{tabular}
}
\label{tab:main_result_coco}\vspace{-3mm}
\end{table*}

\subsection{State-of-the-art Comparison}

Tables~\ref{tab:main_result_pascal} and \ref{tab:main_result_coco} summarize the  comparisons on PASCAL-$5^i$ and COCO-$20^i$, respectively. We evaluate our proposed MK-FSS against recent state-of-the-art non-LLM/MLLM-based and LLM/MLLM-based FSS methods under standard 1-shot and 5-shot settings. As demonstrated, MK-FSS establishes new state-of-the-art performance. Specifically, on PASCAL-$5^i$, it achieves mIoU scores of 87.1\% (1-shot) and 87.7\% (5-shot), surpassing all prior methods. On the more challenging COCO-$20^i$ benchmark, MK-FSS maintains its superiority, delivering highly competitive mIoU scores of 74.3\% and 75.9\% in the 1-shot and 5-shot settings, significantly outperforming existing models. Furthermore, we construct a naive baseline, denoted as MLLM+SAM 2, which directly feeds MLLM-predicted bounding boxes as spatial prompts to SAM 2. The sub-optimal performance of this variant reveals that while MLLMs possess basic detection capabilities, their raw bounding boxes lack the precision strictly required for pixel-level segmentation. This performance gap strongly validates the necessity and effectiveness of our proposed framework.

\subsection{Ablation Study}
To better understand MK-FSS, we conduct ablation studies in the following. Unless explicitly specified, the ablation experiments are conducted on
PASCAL-$5^i$ under the 1-shot setting.

\setlength{\columnsep}{8pt}%
\setlength\intextsep{0pt}
\begin{wraptable}{r}{0.50\textwidth}
\setlength{\tabcolsep}{4pt}
	\centering
	\renewcommand{\arraystretch}{1.1}
    \caption{Ablation on DMDF and PCPG.}
\vspace{-2pt}
\resizebox{\linewidth}{!}{
\begin{tabular}{cccccccc}
\specialrule{1.5pt}{0pt}{0pt}
\rowcolor{mygray}
& DMDF & PCPG & $5^0$ & $5^1$ & $5^2$ & $5^3$ & Mean \\ 
\hline\hline
\ding{182} & - & - & 79.3 & 81.1 & 75.9 & 72.7 & 77.3 \\
\ding{183} & \checkmark & - & 84.8 & 88.2 & 84.1 & 87.0 & 86.0 \\
\ding{184} & - & \checkmark & 80.3 & 82.7 & 77.9 & 77.2 & 80.0 \\
\ding{185} & \checkmark & \checkmark & \textbf{85.0} & \textbf{89.3} & \textbf{86.1} & \textbf{87.8} & \textbf{87.1} \\
\specialrule{1.5pt}{0pt}{0pt}
\end{tabular}
}
\label{tab:component}
 \vspace{1mm}
\end{wraptable}
\textbf{Ablation study on DMDF and PCPG in MK-FSS.} DMDF and PCPG are two crucial modules in MK-FSS for incorporating MLLM-derived knowledge. To better understand them, we conduct ablation experiments in Tab.~\ref{tab:component}. We begin with a model that uses only the support-guided memory feature for FSS (\ding{182}), achieving a mean score of 77.3\%. Using DMDF alone to incorporate MLLM-derived spatial knowledge improves the mean score to 86.0\% (\ding{183}), while PCPG alone for semantic knowledge raises it to 80.0\% (\ding{184}). These results demonstrate the effectiveness of both spatial and semantic knowledge for FSS. When combining DMDF and PCPG, the model achieves the best mean score of 87.1\% (\ding{185}), showing that the two types of knowledge provide complementary benefits and jointly improve FSS.

\textbf{Ablation on different memories.} In MK-FSS, we integrate support-guided memory and MLLM-derived memories through DMDF for segmentation. To analyze the impact of different memories on MK-FSS, we conduct an ablation in Tab.~\ref{tab:memory}. As shown in the table, using either $F_{\text{mem}}^{\text{supp}}$ (\ding{182}) or $F_{\text{mem}}^{\text{MLLM}}$ (\ding{183}) alone leads to suboptimal performance compared to their combination via DMDF, indicating that the two memories provide complementary target guidance. By integrating both memories, MK-FSS effectively leverages their respective strengths and achieves superior performance (\ding{184}).

\textbf{Ablation on stages in DMDF.} DMDF contains three stages of debate, verification, and integration for memory fusion.
To analyze the contribution of each stage in DMDF, we compare the full model (\ding{184}) with partial stages in Tab.~\ref{tab:DMDR}, including a setting using only the debate (Deb.) stage with simple addition for fusion (\ding{182}) and another setting using only the verification and integration (Ver.\&Int.) stages  (\ding{183}). The results show that the three stages collaborate effectively, and combining them is crucial for improving segmentation accuracy.

\textbf{Comparison of DMDF with other fusion strategies.}
To evaluation the effectiveness of our DMDF, we compare it with two alternative strategies in Tab.~\ref{tab:dual-mem fusion}. The first strategy, named learnable addition, combines $F_{\text{mem}}^{\text{supp}}$ and $F_{\text{mem}}^{\text{MLLM}}$ through addition with learnable memory weights (\ding{182}), while the second concatenates two memory features along the channel dimension and fuses them using an MLP (\ding{183}). Compared with these alternatives, our DMDF (\ding{184}) achieves the best performance across all evaluation splits, demonstrating its ability to more effectively exploit the complementary information between support-guided memory and MLLM-derived memory.

\noindent \textbf{Effectiveness of the PCPG Design.}
To evaluate the effectiveness of PCPG, we compare it with two alternative strategies in Tab.~\ref{tab:PCPG}, one relying only on the deepest textual feature $F_{q}^3$ (w/o fusion) (\ding{182}) and the other aggregating multi-scale features $\{F_{q}^i\}_{i=1}^3$ through addition with learnable weights (\ding{183}). Our progressive fusion strategy (\ding{184}) consistently outperforms both the single-level and direct fusion variants, demonstrating that progressively integrating hierarchical features is crucial for constructing a robust multimodal prompt that captures both local details and global semantic cues.

\begin{table*}[!t]
\centering
\setlength{\tabcolsep}{5.5pt}
\begin{minipage}[t]{0.49\textwidth}
\vspace{0pt}
\centering
\renewcommand{\arraystretch}{1.1}
\caption{Ablation on different memories.}
\vspace{-2pt}
\resizebox{\linewidth}{!}{
\begin{tabular}{cccccccc}
\specialrule{1.5pt}{0pt}{0pt}
\rowcolor{mygray}
& $F_{\text{mem}}^{\text{supp}}$ & $F_{\text{mem}}^{\text{MLLM}}$ & $5^0$ & $5^1$ & $5^2$ & $5^3$ & Mean \\ 
\hline\hline
\ding{182} & \checkmark & - & 80.3 & 82.7 & 77.9 & 77.2 & 80.0 \\
\ding{183} & - & \checkmark & 84.1 & 87.8 & 85.9 & 85.2 & 85.8 \\
\ding{184} & \checkmark & \checkmark & \textbf{85.0} & \textbf{89.3} & \textbf{86.1} & \textbf{87.8} & \textbf{87.1} \\
\specialrule{1.5pt}{0pt}{0pt}
\end{tabular}
}
\label{tab:memory}
\end{minipage}
\hfill
\begin{minipage}[t]{0.49\textwidth}
\vspace{0pt}
\centering
\renewcommand{\arraystretch}{1.13}
\caption{Ablation on stages in DMDF.}
\vspace{-2pt}
\resizebox{\linewidth}{!}{
\begin{tabular}{cccccccc}
\specialrule{1.5pt}{0pt}{0pt}
\rowcolor{mygray}
& Deb. & Ver.\& Int. & $5^0$ & $5^1$ & $5^2$ & $5^3$ & Mean \\ 
\hline\hline
\ding{182} & \checkmark & - & 84.7 & 88.8 & 85.7 & 87.0 & 86.6 \\
\ding{183} & - & \checkmark & 84.7 & 88.9 & 85.3 & 87.3 & 86.6 \\
\ding{184} & \checkmark & \checkmark & \textbf{85.0} & \textbf{89.3} & \textbf{86.1} & \textbf{87.8} & \textbf{87.1} \\
\specialrule{1.5pt}{0pt}{0pt}
\end{tabular}
}
\label{tab:DMDR}
\end{minipage}

\vspace{-5pt}
\end{table*}

\begin{table*}[!t]
\centering
\setlength{\tabcolsep}{5pt}
\renewcommand{\arraystretch}{1.1}

\begin{minipage}[t]{0.49\textwidth}
\vspace{0pt}
\centering
\caption{Ablation on the comparison of DMDF with other fusion strategies.}
\vspace{-2pt}
\resizebox{\linewidth}{!}{
\begin{tabular}{ccccccc}
\specialrule{1.5pt}{0pt}{0pt}
\rowcolor{mygray}
& Memory fusion & $5^0$ & $5^1$ & $5^2$ & $5^3$ & Mean \\ 
\hline\hline
\ding{182} & Learnable addition & 84.9 & 88.6 & 85.6 & 86.9 & 86.5 \\
\ding{183} & Concatenation & 84.5 & 88.4 & 85.1 & 86.0 & 86.0 \\
\ding{184} & DMDF & \textbf{85.0} & \textbf{89.3} & \textbf{86.1} & \textbf{87.8} & \textbf{87.1} \\
\specialrule{1.5pt}{0pt}{0pt}
\end{tabular}
}
\label{tab:dual-mem fusion}
\end{minipage}
\hfill
\begin{minipage}[t]{0.49\textwidth}
\vspace{0pt}
\centering
\renewcommand{\arraystretch}{1.13}
\caption{Ablation on fusion method of the
multi-scale cross-modal features of PCPG.}
\vspace{-2pt}
\resizebox{\linewidth}{!}{
\begin{tabular}{ccccccc}
\specialrule{1.5pt}{0pt}{0pt}
\rowcolor{mygray}
& Multi-scale fusion & $5^0$ & $5^1$ & $5^2$ & $5^3$ & Mean \\ 
\hline\hline
\ding{182} & w/o fusion & 84.8 & 88.8 & 85.7 & 86.7 & 86.5 \\
\ding{183} & Learnable addition & \textbf{85.0} & 88.8 & 85.4 & 87.7 & 86.7 \\
\ding{184} & PCPG & \textbf{85.0} & \textbf{89.3} & \textbf{86.1} & \textbf{87.8} & \textbf{87.1} \\
\specialrule{1.5pt}{0pt}{0pt}
\end{tabular}
}
\label{tab:PCPG}
\end{minipage}
\vspace{-5pt}
\end{table*}

\setlength{\columnsep}{8pt}%
\setlength\intextsep{0pt}
\begin{wraptable}{r}{0.50\textwidth}
\setlength{\tabcolsep}{6.5pt}
	\centering
	\renewcommand{\arraystretch}{1.1}
    \caption{Ablation on text embedding.}
\vspace{-4pt}
\resizebox{\linewidth}{!}{
\begin{tabular}{cccccccc}
\specialrule{1.5pt}{0pt}{0pt}
\rowcolor{mygray}
& $F_{s}$ & $F_{w}$ & $5^0$ & $5^1$ & $5^2$ & $5^3$ & Mean \\ 
\hline\hline
\ding{182} & \checkmark & - & 84.8 & 88.2 & 85.4 & 87.4 & 86.4 \\
\ding{183} & - & \checkmark & 84.8 & 88.5 & 85.4 & 87.3 & 86.5 \\
\ding{184} & \checkmark & \checkmark & \textbf{85.0} & \textbf{89.3} & \textbf{86.1} & \textbf{87.8} & \textbf{87.1} \\
\specialrule{1.5pt}{0pt}{0pt}
\end{tabular}
}
\label{tab:Ft}
 \vspace{2pt}
\end{wraptable}
\textbf{Ablation on text embedding.} In MK-FSS, the extracted semantic knowledge is encoded to features that is composed of both sentence embedding $F_s$ with word-level embedding $F_w$. We compare different text embedding strategies in Tab.~\ref{tab:Ft}. The results show that the combination of $F_s$ and $F_w$ achieves superior performance (\ding{184}), validating the efficacy of our design.

Due to limited space, we show more results and analysis in the \emph{\textbf{appendix}}.

\section{Conclusion}
\label{sec:con}
In this paper, we present MK-FSS, a novel framework that exploits both spatial and semantic knowledge derived by MLLMs to enhance FSS through carefully designed DMDF and PCPG modules. Specifically, DMDF incorporates MLLM-derived spatial knowledge into the memory representation by effectively fusing it with the support-guided memory, leading to a more robust dual-memory feature. Meanwhile, PCPG leverages semantic knowledge to enrich visual representations and generate a multimodal prompt that provides additional guidance for target segmentation. By jointly exploiting these two forms of knowledge, MK-FSS captures complementary target cues beyond those available from the limited support samples alone. Extensive experiments on multiple FSS benchmarks demonstrate that MK-FSS consistently achieves strong performance and generalization, validating the effectiveness of incorporating MLLM-derived spatial and semantic knowledge for FSS.

\subsection*{AI use statement}

In this work, the generative AI tools are used \emph{solely} for language polishing and readability improvement. We do \emph{not} use the generative AI to generate datasets, produce experimental results, formulate mathematical claims, design experiments, implement methods, analyze results, or conduct the final interpretation of the findings. All AI-assisted text has been carefully reviewed, revised, and verified by the authors, who take full responsibility for the final content of this work.

\bibliography{iclr2027_conference}
\bibliographystyle{iclr2027_conference}

\newpage

\appendix
\section{Appendix}
The appendix is structured as follows:

\begin{itemize}
\item \textbf{Section~\ref{sec:prompt}}:  We detail the exact prompt templates designed to instruct the MLLMs within our framework and show several examples of the generated spatial and semantic knowledge.

\item 
\textbf{Section~\ref{sec:PGM}}: We present the detailed architecture of the PGM.

\item \textbf{Section~\ref{sec:prior mask}}: Visualization examples of $P_\text{MLLM}$ and $P_\text{supp}$ are provided.

\item \textbf{Section~\ref{sec:error-bar}}: We report the error bar evaluation results under different random seeds to verify the stability and robustness of our model.

\item \textbf{Section~\ref{sec:episodes}}: We analyze the influence of different numbers of testing episodes on model performance on the COCO-$20^i$ dataset.

\item \textbf{Section~\ref{sec:qualitative comparison}}: We present additional qualitative segmentation results and comparisons with other representative models.

\item \textbf{Section~\ref{sec:language_cues}}: We then provide qualitative examples illustrating the effect of linguistic cues, showing how MLLM-derived semantic knowledge enhances visual understanding and disambiguates confusing backgrounds.

\item \textbf{Section~\ref{sec:ca-pcpg}}:  We further examine the effect of cross-attention depth in the PCPG module.

\item \textbf{Section~\ref{sec:MLLM}}:  Additional results evaluated on alternative MLLMs are presented.

\item \textbf{Section~\ref{descriptions}}: We further evaluate the effectiveness of our target-specific language descriptions by comparing them with class-specific descriptions, validating the advantage of our fine-grained semantic information.

\end{itemize}

\subsection{Prompt Design for MLLMs and Examples of Generated Knowledge}
\label{sec:prompt}
\begin{figure*}[htbp]
  \centering
    \centering
    \includegraphics[width=\linewidth]{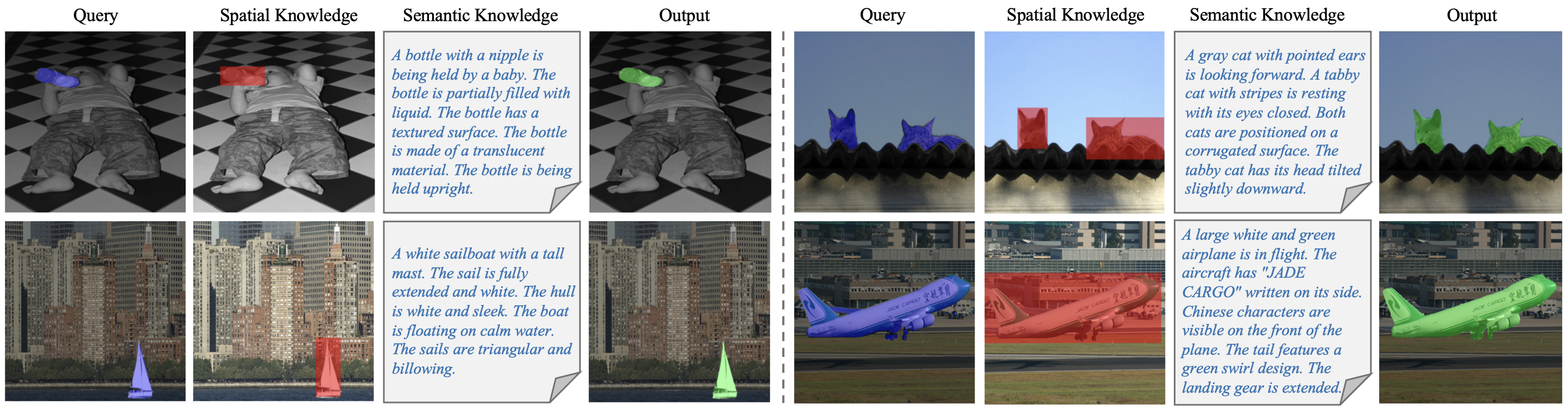}
  \caption{Visualization of MLLM-derived target knowledge and segmentation results on PASCAL-$5^i$ under 1-shot setting. For each example, \textcolor{blue}{blue} masks indicate the ground-truth target regions in query images, \textcolor{red}{red} masks indicate the MLLM-predicted spatial regions, and \textcolor{green}{green} masks indicate the final segmentation results of our proposed MK-FSS.}
    \label{fig:knowledge-examples}  
  \vspace{2mm}
\end{figure*}

We carefully design structured prompts to guide the MLLM in generating precise and task-aligned spatial and semantic knowledge. For spatial knowledge generation, the goal is to obtain structured coordinates that can be directly parsed by downstream components. Specifically, the MLLM is prompted with: \textcolor{promptblue}{\texttt{Locate all the [class\_name] in this tile accurately and completely. Respond ONLY with the bounding box locations in valid JSON format as follows: [{"bbox\_2d": [10, 20, 990, 980], "label": "[class\_name]"}]}}. This prompt encourages the MLLM to localize all potential target instances and return valid bounding boxes. For semantic knowledge generation, we use an explicit restriction prompt to reduce generic semantic noise and fine-grained hallucinations: \textcolor{promptblue}{\texttt{Describe the key visible attributes of all [class\_name] in the image, focusing only on their appearance and state. Use several short factual SENTENCES. Do not mention background objects or irrelevant details. Do not include any guesses about age, condition, history, usage, or purpose}}. These explicit instructions constrain the MLLM to focus exclusively on instance-level visual evidence present in the query scene.

Several examples of the generated spatial and semantic knowledge are shown in Fig.~\ref{fig:knowledge-examples}. We can observe that the two types of knowledge provide complementary target cues. The spatial knowledge roughly localizes potential target regions, while the semantic knowledge describes discriminative visual attributes of the target. By integrating these complementary cues, MK-FSS is able to produce more accurate segmentation results in complicated and challenging scenes.

\subsection{Detailed Architecture of PGM}
\label{sec:PGM}

\begin{wrapfigure}{r}{0.5\textwidth}
\centering
\includegraphics[width=0.5\textwidth]{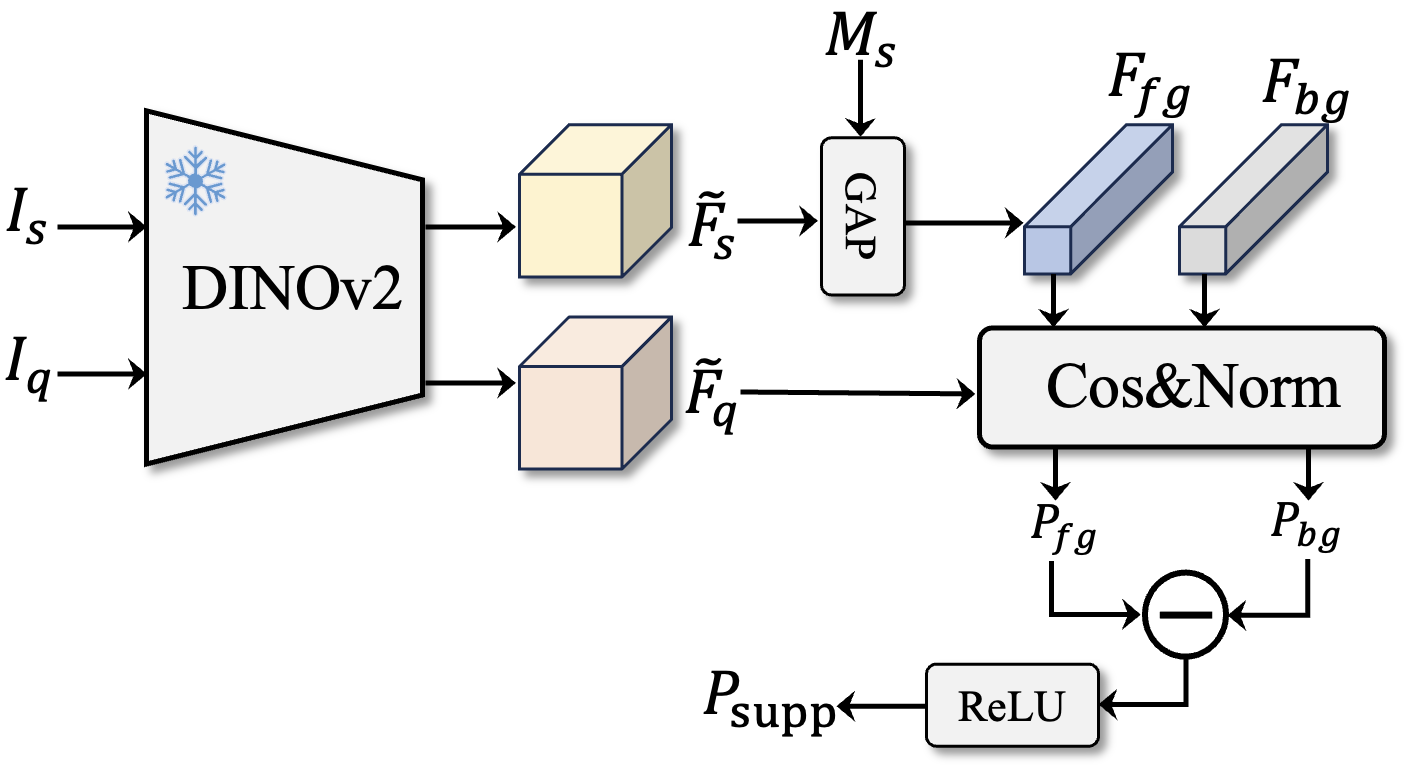}
\caption{Architecture of the PGM.}
\vspace{1mm}
\label{fig:PGM}
\end{wrapfigure}

We illustrate the prior generation module (PGM) in Fig.~\ref{fig:PGM}. PGM generates a support-guided prior map $P_{\text{supp}}$ by measuring the visual correspondence between query features and support prototypes. Following ~\citep{fss-sam}, given a support image $I_s$, a query image $I_q$, and the corresponding support mask $M_s$, we first utilize a frozen DINOv2 encoder~\citep{oquab2023dinov2} to extract the support feature $\tilde{F}_s$ and query feature $\tilde{F}_q$. The foreground and background prototypes are then obtained through the mask-guided global average pooling,
\begin{equation}
F_{fg} = \mathtt{GAP}(\tilde{F}_s \odot M_s), \quad
F_{bg} = \mathtt{GAP}(\tilde{F}_s \odot (1 - M_s))
\end{equation}
We then compute the normalized cosine similarity between the query feature and the two prototypes to obtain the foreground and background similarity maps,
\begin{equation}
P_{fg} = \mathtt{Norm}(\mathtt{CosSim}(\tilde{F}_q, F_{fg})), \quad
P_{bg} = \mathtt{Norm}(\mathtt{CosSim}(\tilde{F}_q, F_{bg}))
\end{equation}
where $\mathtt{Norm}(\cdot)$ denotes min-max normalization. The final support-guided prior is obtained by contrasting the foreground and background similarity maps. Since this operation may introduce negative responses, we apply a ReLU activation to suppress them. This process can be expressed as follows,
\begin{equation}
P_{\text{supp}} = \mathtt{ReLU}(P_{fg} - P_{bg}).
\end{equation}
This contrastive design emphasizes query regions that are more similar to the support foreground while suppressing background-like responses. In the $K$-shot setting, PGM generates one prior map for each support image and averages them to obtain the final $P_{\text{supp}}$.

\subsection{Visualization Examples of $P_\text{MLLM}$ and $P_\text{supp}$}
\label{sec:prior mask}
\begin{wrapfigure}{r}{0.54\textwidth}
  \centering
    \centering
    \includegraphics[width=0.99\linewidth]{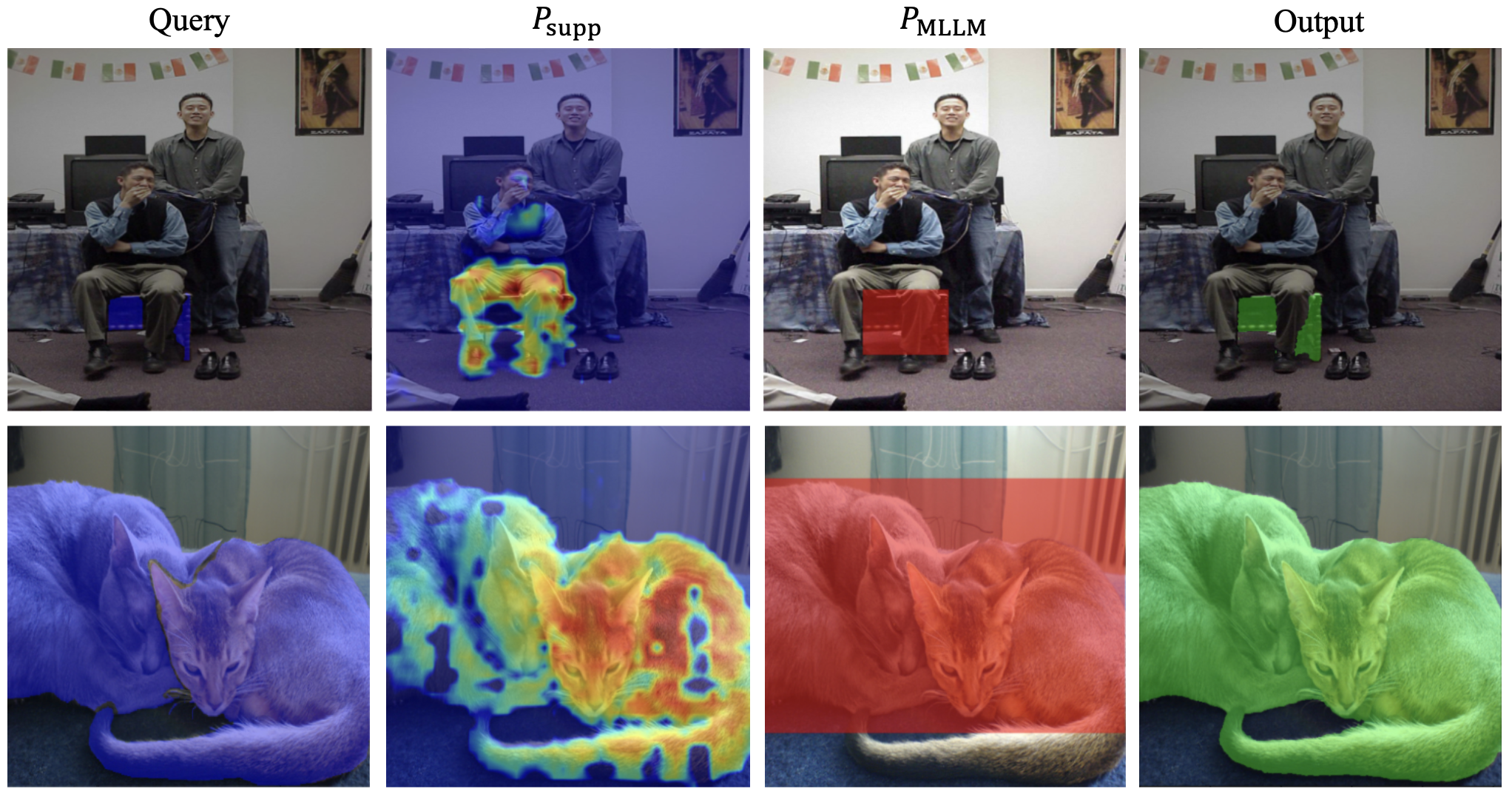}
\caption{Visualization of $P_{\text{supp}}$, $P_{\text{MLLM}}$, and segmentation results on PASCAL-$5^i$. From left to right, each row shows the query image, support-guided prior rendered as a heatmap, MLLM-derived prior, and final prediction. \textcolor{blue}{Blue}, \textcolor{red}{red}, and \textcolor{green}{green} masks indicate ground truth, MLLM-derived prior regions, and predicted masks, respectively.}
    \label{fig:visual-prior}  
  \vspace{-12pt}
\end{wrapfigure}
Fig.~\ref{fig:visual-prior} visualizes the support-guided prior $P_{\text{supp}}$ and the MLLM-derived prior $P_{\text{MLLM}}$. The support-guided prior tends to capture fine-grained visual correspondence, producing detailed activations around the target object. Nevertheless, it may be distracted by visually similar background regions and fail to highlight complete target regions. The MLLM-derived prior provides a more semantic localization cue, which helps identify the target object at a coarse level and suppress irrelevant regions. By combining these two priors, our method benefits from both detailed visual matching and high-level semantic guidance, leading to more complete and accurate segmentation results. These examples qualitatively validate the complementary roles of $P_{\text{supp}}$ and $P_{\text{MLLM}}$.

\subsection{Error Bars Evaluation}
\label{sec:error-bar}
Tab.~\ref{tab:error_bars} shows the error bar evaluation results of our model on the PASCAL-$5^i$ and COCO-$20^i$ datasets under five different random seeds, each tested across four folds with 1,000 episodes. The results demonstrate that our model maintains high stability and robustness across different seeds. Specifically, on the PASCAL-$5^i$ dataset, the mean mIoU remains consistently around 86.8\% with a very small standard deviation of 0.3. And the fold-wise standard deviations are all around 0.5, confirming the model’s consistent generalization ability across different folds. For the more challenging COCO-$20^i$ dataset, although the mean mIoU is relatively lower (73.3\%), the standard deviation remains small, showing stable performance despite higher data complexity. The fold-wise variations are also limited, with Fold-1 achieving the best results and Fold-0 showing slightly larger variance, which stems from higher intra-class diversity of COCO-$20^i$. Overall, the low standard deviations across both datasets validate the reliability of our MK-FSS and its robustness to random sampling.

\begin{table*}[t]
\setlength{\tabcolsep}{9pt}
\centering
\caption{Error bars evaluation on PASCAL-$5^i$ and COCO-$20^i$ datasets under 1-shot setting. The number of testing episodes is 1,000. The selected random seeds are \{0, 1, 2, 3, 321 (default)\}. “Mean” is the averaged mIoU score of 4 folds.}
\resizebox{0.88\textwidth}{!}{  
\begin{tabular}{rcccccccccc}
\specialrule{1.5pt}{0pt}{0pt}
\multirow{2}{*}{\textbf{Seed}} & \multicolumn{5}{c}{\textbf{PASCAL-$5^i$}} & \multicolumn{5}{c}{\textbf{COCO-$20^i$}} \\
\cmidrule(lr){2-6} \cmidrule(lr){7-11}
& $\mathbf{5^0}$ & $\mathbf{5^1}$ & $\mathbf{5^2}$ & $\mathbf{5^3}$ & \textbf{Mean} & $\mathbf{20^0}$ & $\mathbf{20^1}$ & $\mathbf{20^2}$ & $\mathbf{20^3}$ & \textbf{Mean} \\
\hline\hline
0 & 84.8 & 89.0 & 85.3 & 87.2 & 86.6 & 71.9 & 77.5 & 73.8 & 69.4 & 73.2\\
1 & 85.0 & 89.2 & 86.2 & 87.7 & 87.0 & 69.1 & 78.4 & 73.0 & 73.3 & 73.5\\
2 & 84.3 & 88.5 & 84.4 & 87.8 & 86.3 & 66.6 & 76.3 & 73.9 & 72.7 & 72.4\\
3 & 84.3 & 89.1 & 86.3 & 87.5 & 86.8 & 71.9 & 75.5 & 70.9 & 73.9 & 73.1 \\
321 & 85.0 & 89.3 & 86.1 & 87.8 & 87.1 & 73.2 & 76.6 & 75.9 & 71.6 & 74.3\\
\rowcolor[HTML]{eaf4fc}
Mean & 84.7 & 89.0 & 85.7 & 87.6 & 86.8 &  70.5 & 76.8 & 73.5 & 72.2 & 73.3\\
\rowcolor[HTML]{EFF7E6} 
Std & 0.4 & 0.3 & 0.8 & 0.3 & 0.3 & 2.7 & 1.1 & 1.8 & 1.8 & 0.7\\
\specialrule{1.5pt}{0pt}{0pt}
\end{tabular}
}
\label{tab:error_bars}\vspace{-0mm}
\end{table*}

\subsection{Different Number of Testing Episodes}
\label{sec:episodes}

\begin{wraptable}{r}{0.5\textwidth}
\centering
\setlength{\tabcolsep}{6pt}
\renewcommand{\arraystretch}{1.12}
\caption{Performance on COCO-$20^i$ with different number of testing episodes in \{1,000, 4,000, 10,000, 20,000\}. The random seed is fixed as 321. “Mean” is the averaged mIoU score of 4 folds, “$20^i$” denotes the mIoU score of the i-th fold.}
\resizebox{0.5\textwidth}{!}{
\begin{tabular}{cccccc}
\specialrule{1.5pt}{0pt}{0pt}
\textbf{Testing Episodes} & $\mathbf{20^0}$ & $\mathbf{20^1}$ & $\mathbf{20^2}$ & $\mathbf{20^3}$ & \textbf{Mean} \\
\hline\hline
1,000  & 73.2   & 76.6   & 75.9   & 71.6   & 74.3 \\
4,000  & 72.7   & 75.5   & 72.2   & 72.5   & 73.2 \\
10,000 & 71.2   & 75.4   & 73.4   & 72.7   & 73.2 \\
20,000 & 70.6   & 75.7   & 73.2   & 72.6   & 73.0  \\
\rowcolor[HTML]{EFF7E6} 
Mean   & 71.9   & 75.8   & 73.7   & 72.4   & 73.5  \\
\specialrule{1.5pt}{0pt}{0pt}
\end{tabular}
}
\label{tab:test_episodes}
\vspace{-2mm}
\end{wraptable}
Since COCO-$20^i$ is a challenging dataset, the evaluation results based on only 1,000 randomly sampled testing episodes may not be sufficiently convincing. Following ~\citep{fss-sam}, we further investigate the influence of different testing episode numbers on model performance. Specifically, we vary the number of testing episodes among {1,000, 4,000, 10,000, 20,000} under 1-shot setting and report the corresponding quantitative results in Tab.~\ref{tab:test_episodes}.

From the results, it can be observed that varying the number of testing episodes has a slight impact on the performance. For example, the mIoU achieved with 1,000 testing episodes is 74.3\%, while the average mIoU across all four episode settings is 73.5\%, yielding a marginal gap of only 0.8\%. This demonstrates that our model achieves stable and reliable performance even with a relatively small number of testing episodes. Furthermore, the fold-wise variations remain consistent across different episode settings, indicating that the model generalizes well regardless of the number of sampled episodes. Overall, these results validate the robustness and consistency of our approach when evaluated under different testing conditions.

\subsection{Qualitative Analysis}
\label{sec:qualitative comparison}

\begin{figure*}[htbp]
  \centering
    \centering
    \includegraphics[width=\linewidth]{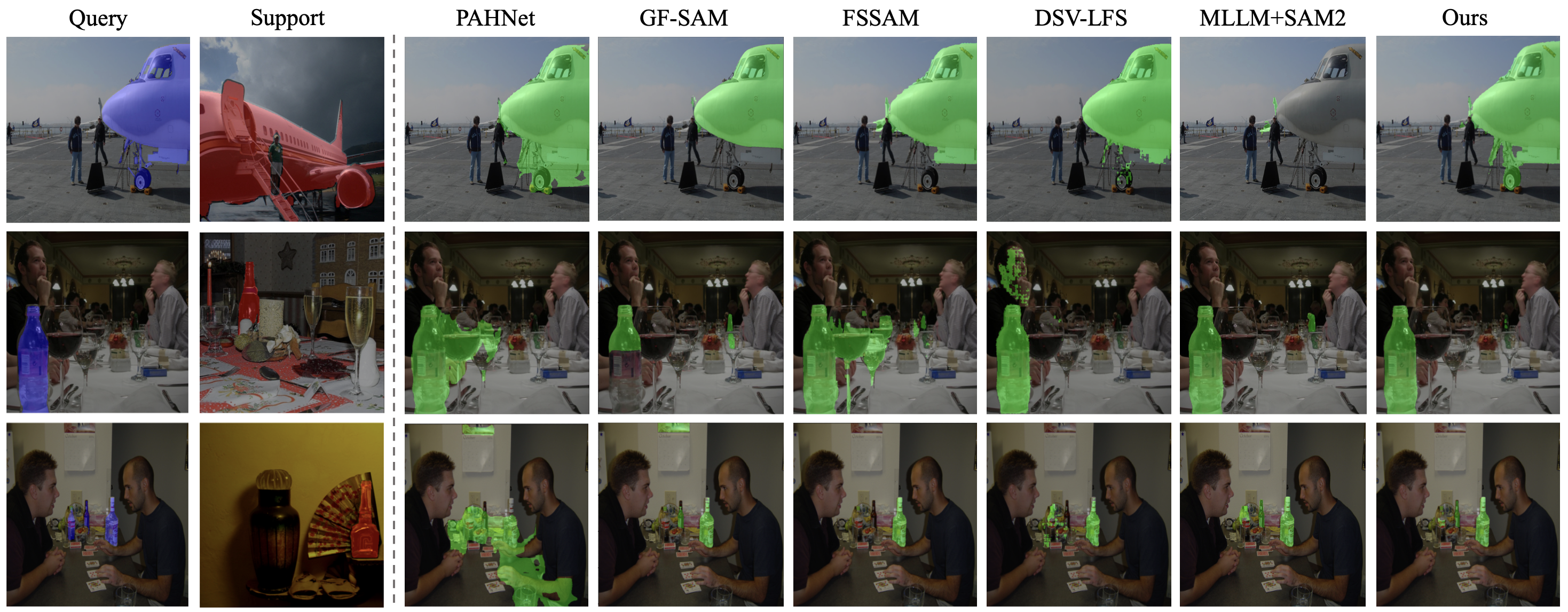}
  \caption{Visual comparison on PASCAL-$5^i$ dataset under 1-shot setting. The ground-truth objects in query image are highlighted in \textcolor{blue}{blue}, the referred objects in support image are highlighted in \textcolor{red}{red}, and the predicted objects in query image are highlighted in \textcolor{green}{green}.}
    \label{fig:visual-compare}  
  \vspace{-12pt}
\end{figure*}

To further qualitatively validate the effectiveness of our proposed method, we provide visual segmentation results and compare them with representative methods~\citep{pahnet,gf-sam,fss-sam,ravi2024sam,dsv-lfs}, as shown in Fig.~\ref{fig:visual-compare}. It can be observed that existing methods often produce incomplete masks or include irrelevant background regions, especially in challenging cases with occlusions and visually similar distractors, e.g., the bottles in Fig.~\ref{fig:visual-compare}. In contrast, our method generates more accurate and complete target masks with clearer object boundaries. These results demonstrate that MK-FSS effectively exploits MLLM-derived spatial and semantic knowledge, which provides complementary localization and semantic cues to enhance the model's segmentation capability in complex scenarios.

\subsection{Visualization of the Effect of Linguistic Cues}
\label{sec:language_cues}
\begin{figure*}[ht]
  \centering
    \includegraphics[width=0.9\linewidth]{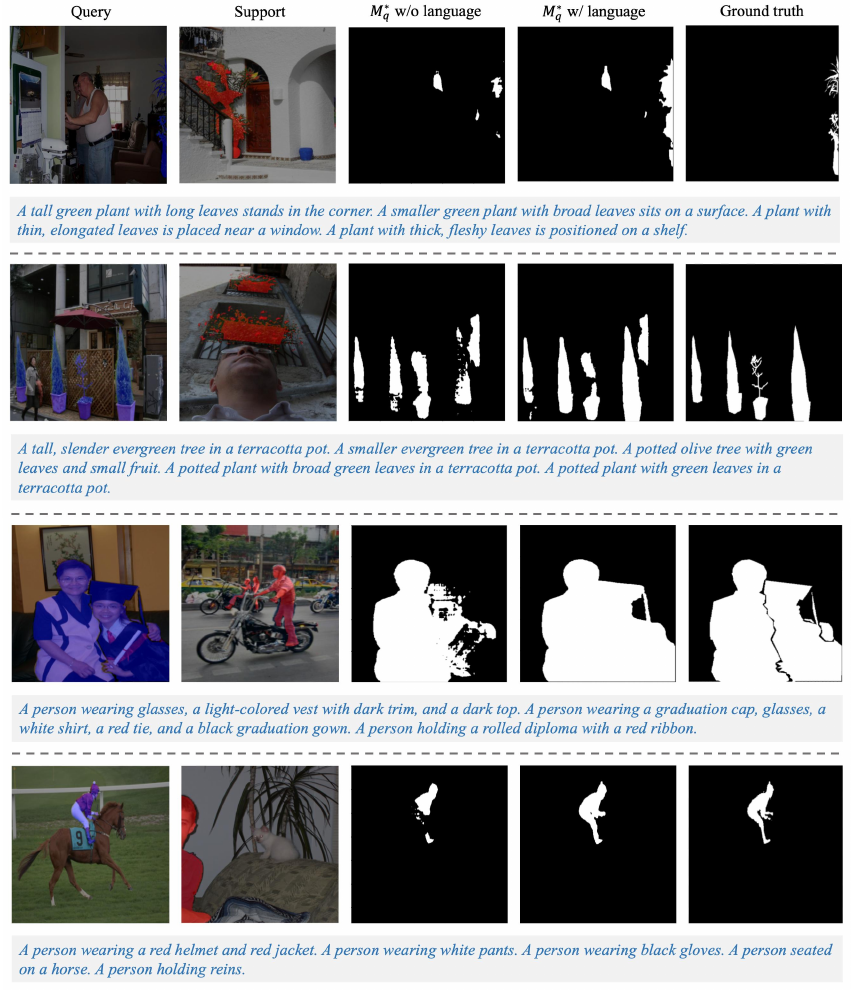}
  \caption{Qualitative examples showing the effect of linguistic cues on visual understanding.}
  \label{fig:language_cues}
  \vspace{-3mm}
\end{figure*}

In Fig.~\ref{fig:language_cues}, we present several examples illustrating the impact of incorporating high-level linguistic cues into purely visual  features. From these examples, we can see that models relying solely on visual matching often struggle to handle complex scenarios, where the same object category shows significant variations in appearance across different images (e.g., the people and plant in Fig.~\ref{fig:language_cues}). However, the integration of linguistic information provides additional semantic priors that offer explicit and discriminative guidance for the visual features. Such language-guided cues help the model focus on the semantically relevant regions and disambiguate visually similar background objects, leading to more accurate and robust segmentation results.
\label{sec:B}

\subsection{Impact of Cross-Attention Depth in PCPG}
\label{sec:ca-pcpg}

\begin{wraptable}[7]{r}{0.48\textwidth}
\vspace{0pt}
\centering
\setlength{\tabcolsep}{4pt}
\renewcommand{\arraystretch}{1.08}
\caption{Ablation on CA number in PCPG.}
\vspace{-6pt}
\resizebox{\linewidth}{!}{
\begin{tabular}{ccccccc}
\specialrule{1.5pt}{0pt}{0pt}
\rowcolor{mygray}
& CA layer number & $5^0$ & $5^1$ & $5^2$ & $5^3$ & Mean \\
\hline\hline
\ding{182} & 1 & 84.8 & 88.8 & 85.8 & 87.7 & 86.8 \\
\ding{183} & 2 & \textbf{85.0} & \textbf{89.3} & \textbf{86.1} & \textbf{87.8} & \textbf{87.1} \\
\ding{184} & 3 & 84.9 & 88.8 & 85.5 & 87.5 & 86.7 \\
\specialrule{1.5pt}{0pt}{0pt}
\end{tabular}
}
\label{tab:layer-num}
\vspace{-10pt}
\end{wraptable}
To examine the effect of cross-attention depth in the PCPG module, we vary the number of cross-attention layers from 1 to 3. The quantitative results, presented in Tab.~\ref{tab:layer-num}, indicate that the 2-layer design (\ding{183}) yields the optimal balance and achieves the highest overall performance.

\subsection{Analysis of Impact across MLLMs}
\label{sec:MLLM}
To show that the performance gains are brought by the proposed architecture rather than tied to a specific MLLM, we evaluate several recent MLLMs within our framework. As shown in Tab.~\ref{tab:MLLM}, our method consistently achieves strong performance across different MLLM choices. Notably, using a more advanced model, Qwen3-VL-8B (\ding{186}), further improves the mean mIoU to 87.1\%. These results indicate that our framework can effectively leverage target knowledge from different MLLMs, and its performance can further benefit from the improved reasoning capabilities of newer models.
 \vspace{1em}
\begin{table}[htbp]
\centering
\setlength{\tabcolsep}{4pt}
\renewcommand{\arraystretch}{1.1}
\caption{Performance comparison under different MLLM configurations on PASCAL-$5^i$.}
\vspace{-4pt}
\resizebox{0.83\linewidth}{!}{
\begin{tabular}{ccccccc}
\specialrule{1.5pt}{0pt}{0pt}
\rowcolor{mygray}
& MLLM & $5^0$ & $5^1$ & $5^2$ & $5^3$ & Mean \\
\hline\hline
\ding{182} & Qwen2.5-VL-7B-Instruct~\citep{Qwen2.5-VL}  & 83.7 & 89.2 & 81.7 & 85.0 & 84.9 \\
\ding{183} & Qwen2.5-VL-32B-Instruct~\citep{Qwen2.5-VL} & 82.8 & 89.7 & 83.8 & 87.5 & 86.0 \\
\ding{184} & MiMo-VL-7B-RL~\citep{mimo} & \textbf{85.9} & \textbf{90.2} & 85.2 & 86.3 & 86.9 \\
\ding{185} & InternVL3.5-8B~\citep{internvl3_5} & 83.0 & 87.8 & 84.8 & 85.8 & 85.4 \\
\ding{186} & Qwen3-VL-8B-Instruct~\citep{qwen3-vl} & 85.0 & 89.3 & \textbf{86.1} & \textbf{87.8} & \textbf{87.1} \\
\specialrule{1.5pt}{0pt}{0pt}
\end{tabular}
}
\label{tab:MLLM}
\vspace{-5pt}
\end{table}

\vspace{1em}

\subsection{Effectiveness of Our Target-specific Language Description}
\label{descriptions}
\begin{wraptable}[6]{r}{0.55\textwidth}
\centering
\setlength{\tabcolsep}{6pt}
\renewcommand{\arraystretch}{1.15}
\caption{Performance comparison under different language description configurations.}
\vspace{-4pt}
\resizebox{1\linewidth}{!}{
\begin{tabular}{ccccccc}
\specialrule{1.5pt}{0pt}{0pt}
\rowcolor{mygray}
& Language Description & $5^0$ & $5^1$ & $5^2$ & $5^3$ & Mean \\
\hline\hline
\ding{182} & Class-specific  & 84.7
 & 89.2
 & 84.4
 & 87.2
 & 86.4 \\
\ding{183} & Target-specific &\textbf{85.0} & \textbf{89.3} & \textbf{86.1} & \textbf{87.8} & \textbf{87.1} \\
\specialrule{1.5pt}{0pt}{0pt}
\end{tabular}
}
\label{tab:language-use}
\vspace{-5pt}
\end{wraptable}
To further verify the effectiveness of our generated target-specific language descriptions, we conduct a comparative experiment between using two types of textual descriptions: \ding{182} the class-specific descriptions generated by ChatGPT-4.0 as adopted in DSV-LFS~\citep{dsv-lfs}, and \ding{183} our proposed target-specific descriptions, which are generated individually for each query image. As shown in Tab.~\ref{tab:language-use}, our image-specific descriptions achieve consistently higher performance across all folds on PASCAL-$5^i$. This improvement demonstrates that general class-level language descriptions are insufficient to capture fine-grained and context-dependent visual details, while our specific descriptions provide more precise semantic cues that better align with the specific visual content of each image. 

\end{document}